\pdfoutput=1
\documentclass{article}
\usepackage{ijcai22}

\usepackage{times}
\usepackage{soul}
\usepackage{url}
\usepackage[hidelinks]{hyperref}
\usepackage[utf8]{inputenc}
\usepackage[small]{caption}
\usepackage{graphicx}
\usepackage{amsmath}
\usepackage{amsthm}
\usepackage{booktabs}
\usepackage{algorithm}
\usepackage{algorithmic}
\usepackage{xcolor}

\definecolor{myblue}{HTML}{3A87FE}
\definecolor{myorange}{HTML}{FECB3E}
\definecolor{mygreen}{HTML}{76BB40}

\title{From Correlation to Causation:\\ Formalizing Interpretable Machine Learning as a Statistical Process}

\author{
Lukas Klein$^{1,2,3}$
\and
Mennatallah El-Assady$^3$\And
Paul Jäger$^{1,2}$
\affiliations
$^1$Interactive Machine Learning Group, DKFZ, Germany\\
$^2$Helmholtz Imaging, DKFZ, Germany\\
$^3$Department of Computer Science, ETH Zürich, Switzerland
\emails
\{lukas.klein, p.jaeger\}@dkfz.de,
menna.elassady@ai.ethz.ch
}

\begin{document}

\maketitle

\begin{abstract}
    Explainable AI (XAI) is a necessity in safety-critical systems such as in clinical diagnostics due to a high risk for fatal decisions. Currently, however, XAI resembles a loose collection of methods rather than a well-defined process. In this work, we elaborate on conceptual similarities between the largest subgroup of XAI, interpretable machine learning (IML), and classical statistics. Based on these similarities, we present a formalization of IML along the lines of a statistical process. Adopting this statistical view allows us to interpret machine learning models and IML methods as sophisticated statistical tools. Based on this interpretation, we infer three key questions, which we identify as crucial for the success and adoption of IML in safety-critical settings. By formulating these questions, we further aim to spark a discussion about what distinguishes IML from classical statistics and what our perspective implies for the future of the field.  
\end{abstract}

\section{Introduction}

\begin{figure*}[!ht]
    \centering
    \includegraphics[trim=0cm 4.7cm 0cm 9cm,clip, width=\linewidth, page = 1]{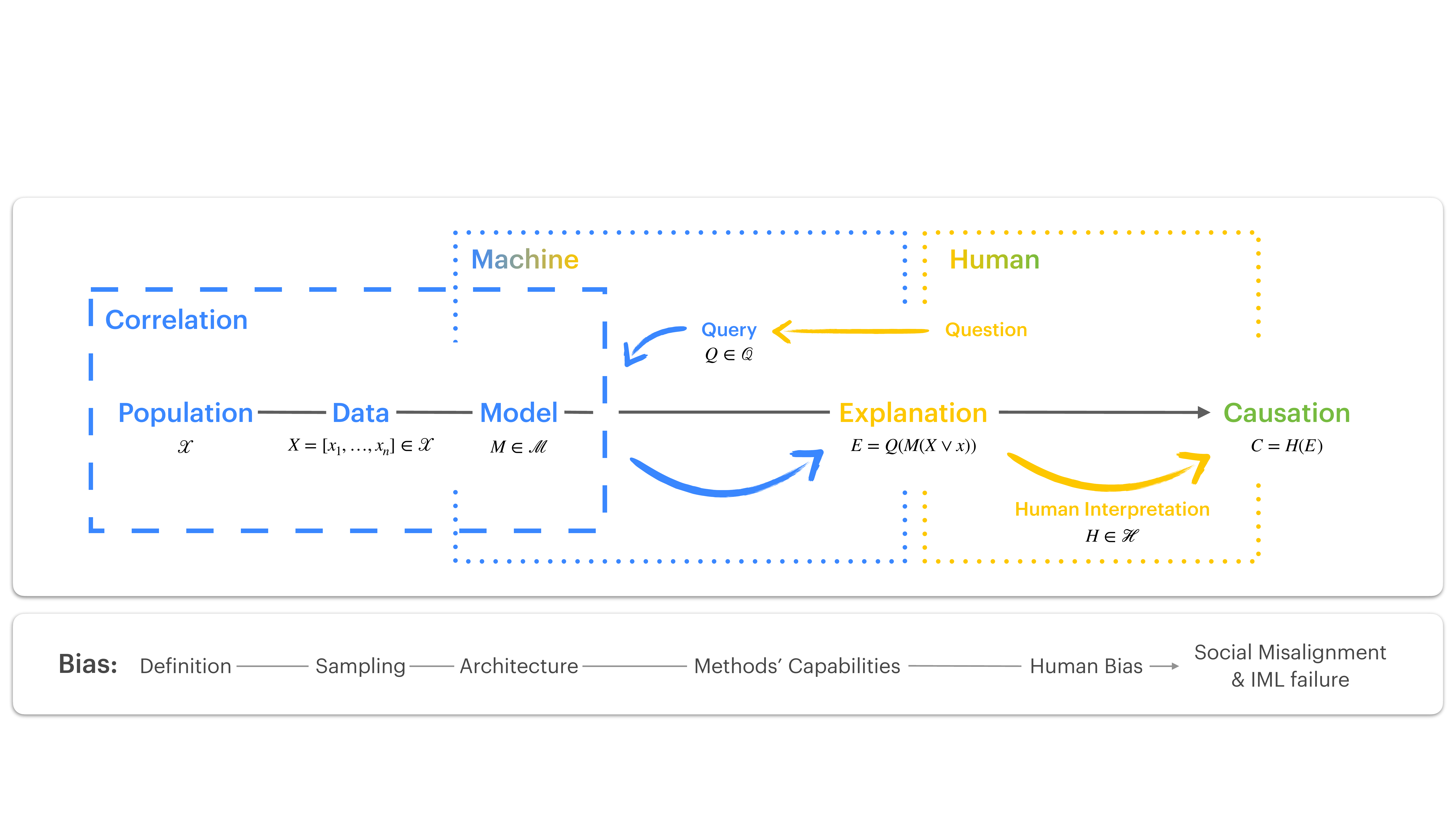}
    \caption{\footnotesize The IML process leading from correlation to causation. The explanation connects the machine with the human to leverage their synergies and individual strengths. For instance, while machines can capture and summarize a large amount of information, humans are better at generalizing this information. Further, at every stage in the process, a new inductive bias is introduced, not limiting to the ones listed at the bottom.}
    \label{fig1}
\end{figure*}

Artificial intelligence (AI) has to be trustworthy to succeed not only in real-world applications but also in safety-critical settings such as clinical diagnostics. From the engineering perspective (ignoring social and psychological factors), this trust is either established by showing exceptionally robust performance and succeeding under diverse conditions, or by providing transparent decision-making and explaining precisely when and why a model failed or prevailed \citep{mackowiak_generative_2020,chen_interpretable_2021}. While initially, robustness might sound like the more appealing option, it is difficult to achieve and verify  \citep{Survey_Huang_2018,practical_Mohseni_2019}. This is of particular importance in medical applications where models are employed to support the diagnostic process of the physician, which, however, can neither adequately verify the model's predictions nor compare its decision process to their own. Explainable AI (XAI), on the other hand, allows for specific interactions with different stakeholders in a model's life-cycle such as, e.g., debugging and robustness for engineers, interaction and understanding for domain-experts, and fairness and data protection for regulators \citep{holzinger_towards_2017,holzinger_survey_2020,chen_interpretable_2021}. In the case of regulators they are not only requesting explainability but moving in the direction of demanding it, and making concerns of social interest, such as ethics and safety, legally binding \citep{Selbst2017,Wachter2017,european-commission-2021}. Ultimately, both approaches are coexisting, but XAI should not be depreciated by arguing that models only have to be robust.\\

Currently, however, XAI lacks a clear formalization as it resembles more of a loose collection of methods, for which again it is not clearly defined when a method is \textit{explainable}. The largest subgroup of XAI is interpretable machine learning (IML), concerned with the explanation of machine and deep learning models. Other subgroups include, e.g., interpretable reinforcement learning \citep{NEURIPS2019_e9510081} or interpretable evolutionary algorithms \citep{Leguy2020}. In this work, we argue that one possible formalization of IML can be by means of statistics, indicating that we might have been doing classical statistics all along. First, we define IML as a process leading from correlation to causation. Furthermore, we elaborate on conceptual similarities between IML and classical statistics and infer a formalization of IML along the lines of a statistical process. We argue, that through our formalization, IML can be defined as a statistical process leveraging machine learning models and query methods, viewing IML methods as sophisticated statistical tools to support or contradict hypotheses. At last, we present three key questions elaborated from this view, which we argue are crucial for the success and adoption of IML in real-world and safety-critical settings.

\section{The Formalization of IML} \label{sec:DefIML}

We are formalizing the problem of IML not only as a repertoire of methods but by proposing a chain-like process of procedures, inspired by the notation of \citet{barcelo_model_2020}. While direct modelling of causal relations is still in a preliminary stage \citep{scholkopf2019causality}, IML methods are applied to render correlation human-understandable as an intermediary step via either inherently interpretable models or post-training explanatory methods \citep{holzinger_toward_2021,marcinkevics_interpretability_2020}. \autoref{fig1} shows how the human $H$ from the class of stakeholders, $\mathcal{H}$ then is interpreting and utilizing this machine-based explanation $E$ to make a  causal statement $C$. $C$ can incorporate causal statements about, e.g., fairness, reliability, or transparency, which also can be wrong.\\

\begin{figure*}[!ht]
    \centering
    \includegraphics[trim=2cm 11cm 2cm 11cm,clip, width=\linewidth, page = 2]{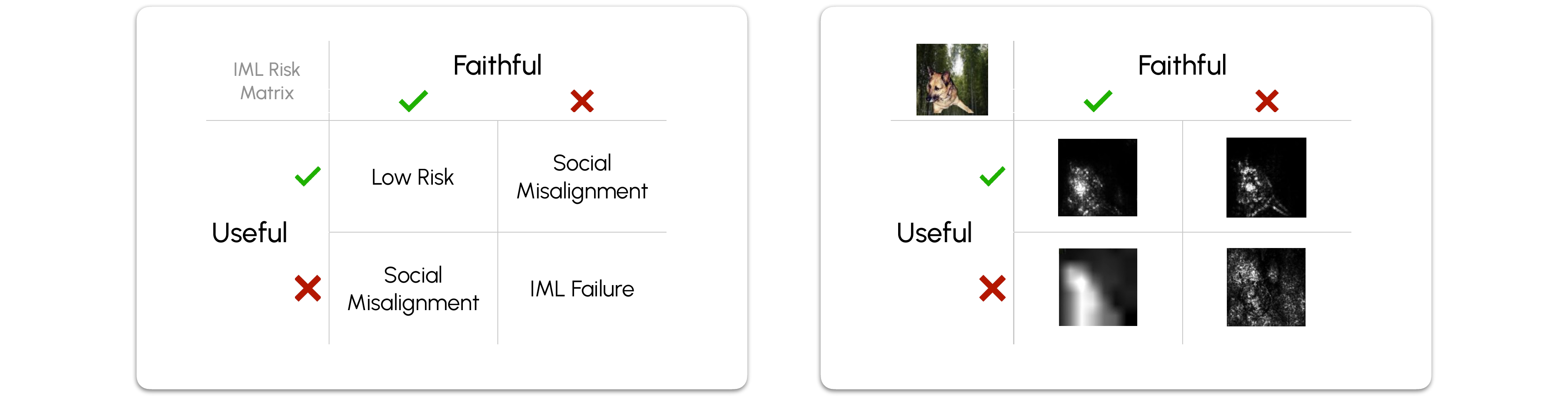}
    \caption{\footnotesize \textbf{Left:} If an explanation is faith-and useful, the risk of making a wrong interpretation is rather low. But when an explanation is largely unfaithful, the risk of taking a wrong explanation as true increases. Similar a largely unuseful, but true, explanations can easily be over-or misinterpreted, resulting again in a wrong explanation. Both are forms of social misalignment and are continuously measured since it is very hard, if not impossible, to discretely measure them. If the query fails entirely to retrieve the \textit{true} explanation and to communicate the resulting explanations, one speaks of IML failure. \textbf{Right:} Examples based on attribution methods.}
    \label{fig2}
\end{figure*}

We summarize the process from data definition to distilled relationship as correlation, since it describes the creation and transformation of the relationships as a raw mathematical object, still uninformative to the human. The basis of correlation is determined by the unobserved population $\mathcal{X}$, sometimes called ground truth in machine learning (not to confuse with ground truth label), from where we sample a dataset $X$. Further, we induce a model $M$ of the model-class $\mathcal{M}$ to extract correlation from the dataset. Depending on our question we want to make a causal statement about, we have to design a query $Q$ from a query-class $\mathcal{Q}$, which describes the interpretable mechanism we either apply as a method upon our model $M$ or implement as an inductive bias into it. $Q$ acts either on a single instance $x \in X$ for a local explanation or on $X$ for a global. In application, $\mathcal{Q}$ is defined as a subgroup of interpretable models or explanatory methods solving for the same question, such as, e.g., attribution methods solving for feature importance \citep{sundararajan_axiomatic_2017,Lundberg_2017_Unified,linardatos_explainable_2020}. $Q$ is then defined as the specific method, e.g., Integrated Gradients \citep{sundararajan_axiomatic_2017}. The design of the query is crucial in the sense that it has to be faithful to the model \emph{and} useful (sometimes also called comprehensible) to the human \citep{chen_interpretable_2021}. Otherwise, the induced bias by the query can lead to social misalignment \citep{jacovi_aligning_2021} (see \autoref{fig2}, right images from \citet{yang_benchmarking_2019}), \textit{``[\dots] a situation where an explanation communicates a different kind of information than the kind that people expect it to communicate.''} \citep{hase_exploring_2021}. \\

But not solely ill-defined queries that bias the explanation can ultimately lead to a \textit{false} causality statement. At every stage in the process, a bias is induced (\autoref{fig1}, bottom), limiting the power of the causal statement, and increasing the risk for social misalignment or even IML failure. The definition of a faithful, thus \textit{true} explanation, is non-trivial since there is no given ground truth label and it depends on the underlying model. While some biases are induced by design such as sampling procedure, model architecture, or query method selection, others are more unconscious such as various human biases. However, these biases do not necessarily prevent us from reaching a \textit{true} causal statement \citep{doshi-velez_towards_2017}. In fact, with statistics, there is a whole scientific discipline which concern is to organize, analyze, interpret, and present model and data utilizing limited methods inducing bias to reach a \textit{true} causal statement. 

\section{The State of IML}

To connect the abstract definition of a query and explanation to concrete IML methods, we outline the state-of-the-art in the field and the direction it is progressing. We ignore methods used for sampling data or estimating models since they are beyond the scope of this work. The selection of the IML method fulfilling the query $Q$ is one of the most crucial steps in the process. We structure the query-class $\mathcal{Q}$ into interpretable models and explanatory methods. Interpretable models are of cause deeply connected to the selection of the model-class $\mathcal{M}$ but describe the inherit interpretable mechanism of the model, detached from other architectural designs. The most prominent class of interpretable models are classical statistical models such as generalized, mixed, additive, or simpler linear models \citep{mccullagh1989generalized,McCulloch2001}. Neural additive models \citep{agarwal_neural_2021} are an extension of additive models, making them capable to be applied to more complex problems where they otherwise lack predictive performance. Other machine learning-based interpretable models include interpretable representations (disentangled or factorized representations) \citep{higgins_beta-vae_2017,chen_isolating_2019,kim_disentangling_2019}, sparse (input) models \citep{Feng_2017_Sparse,Tibshirani94regressionshrinkage}, or graph models (causal graphs or various tree-based methods) \citep{Glymour_2019_causalgraph,scholkopf_towards_2021,Bishop_2006_Pattern,hastie2009elements}. Proxy interpretable models, which do not have the main objective to be interpretable but still have an interpretable component, include, e.g., attention-based \citep{NIPS2017_3f5ee243} and probabilistic models \citep{Murphy_2021_Prob}. Probabilistic models allow for interpretative statements about generative processes and uncertainties.\\

Interpretable models were the default methods in IML for a long time, but with the advent of deep learning the focus shifted to explanatory methods, allowing for more complex models without performance reducing assumptions imposed by many interpretable models. Explanatory methods are post-training applied to the model and can be clustered into model agnostic vs specific and global vs local methods \citep{marcinkevics_interpretability_2020}. Momentarily, attribution methods are the preferred query-class $\mathcal{Q}$ by the field for explanatory methods. They measure the importance of a local feature (e.g. pixels in an image or words in a sentence) into an output by means of gradients \citep{sundararajan_axiomatic_2017,Selvaraju_2017_gradcam,Springenberg_2014_guidedback,Bach_2015_LRP}, Shapley values \citep{Lundberg_2017_Unified}, perturbations \citep{ZeilerF13,Ribeiro_2016_LIME}, or latent features \citep{Schulz2020Restricting}. Other query-classes include, e.g., counterfactual explanations \citep{wachter2018a}, symbolic meta-models \citep{NEURIPS2019_567b8f5f}, or search methods \citep{Ribeiro_Singh_Guestrin_2018}. Recently, concept-based methods were introduced, testing if a model learned a certain concept which is derived by either human (CAV) \citep{kim_interpretability_2018} or machine (ACE) \citep{ghorbani_towards_2019}, and allowing for more global explanations in computer vision. Likewise, explanatory methods contain proxy methods such as adversarial examples \citep{Goodfellow_2015_adver}, which can be characterized as a subgroup of counterfactual examples.\\

Recently, the trend is moving in the direction of applying queries in a more human-centred convention through designed application frameworks \citep{holzinger_toward_2021,liao_human-centered_2022,chen_interpretable_2021} and interactive tools \citep{Menna_2019_expl,Baniecki_2021_Dalex}, compared to the algorithm-centred application. These frameworks define the selection and evaluation of the different IML methods not only by technical but also by use-case-specific means. They guide the human stakeholder $H$ to select a suitable query $Q$ for their question and bridge the gap between method-based explanation and actionable conclusion. Besides technical maps, these frameworks often leverage psychological-based behavioural templates to minimize the possibility of miscommunication, e.g., between physician and clinical decision system \citep{Jussupow_2021} and define how $H$ can optimally leverage an explanation $E$ to achieve a causal statement $C$. Interactive tools utilize the fact that different representations (visual, verbal, etc.) of the same explanation can increase and evaluate the usefulness to different stakeholders. The evaluation of queries is primarily performed through metrics to evaluate faithfulness \citep{yeh_fidelity_2019} and human-based experiments to evaluate usefulness \citep{wang_are_2021,holzinger_survey_2020}.\\

However, there is still a gap between research and practice, hindering the adoption of IML, with most current critics being methodological. For example, it was shown that several approaches are not robust or consistent in their explanation \citep{NEURIPS2018_294a8ed2,shen2019evaluating,pmlr-v80-nie18a} and point out untrue relationships in the data or model, e.g., rather reconstructing the input image or performing edge detection than estimating the feature importance \citep{NEURIPS2018_294a8ed2}. Further, untrue explanations are often accepted through the human confirmation bias, falsely rejecting true but less human appealing explanations \citep{yang_benchmarking_2019}. Evaluation methods lack a ground truth label and, as a proxy, use performance metrics such as accuracy to determine the importance of a feature to the model's prediction \citep{yeh_fidelity_2019}. This results in an evaluation of correlation rather than causal concepts \citep{wang_are_2021}. Further, the evaluation through removing attributed features can create out-of-distribution (OOD) or adversarial examples, which then evaluate more the robustness of the model than the importance of the features \citep{hase_exploring_2021}. While methodological criticism is important to the engineering of IML methods, the focus of our proposal lies on the formal process, which either adapts current methods to its needs or highlights where the repertoire has to be expanded. \\


\section{Comparing Statistical Formalism to IML} \label{sec:StatForm}

The higher maxim behind the current progress in IML tries to satisfy ill-defined problems like trust or causality, which are extremely hard if at all possible to achieve by proposing single methods. We argue that the focus of the progress has to shift, developing, applying, and evaluating queries based on less glorious objectives. IML methods should be handled as statistical tools, enriching hypotheses by potential evidence for or against them.\\

The proposition to interpret IML methods more as statistical diagnosis tools was already mentioned by \citet{chen_interpretable_2021}, but they make no more far-reaching statement about how to accomplish this or what the implications would be. While proposed frameworks work great for method selection and cover the practical procedure of experiment planning, they do not offer a general formalism in which they are embedded. Formalisms in science describe the symbolic manner in which information is presented \citep{sep-formalism-mathematics}. Statistical formalism differs in the sense that it does not infer its formal system based on axioms, but based on a standardized procedure. This procedure can be summarized by means of sampling, modelling, and testing. See \citet{romeijn_2014} for a detailed definition. Sampling is defined as the selection of a subset of observations from the population which yields optimally the same characteristics as the whole population. Modelling describes the practice of either descriptive analysis, summarizing and making statements about the sample data, or inference, drawing conclusion about causality which hold in general also for the population. When interpreted in a broader spectrum, testing includes the definition of hypotheses, error types, confidence, and significance. Testing is important to counter the miss-use and miss-interpretation of statistics, especially when using ill-defined methods, inducing bias.\\

First, we compare the process of statistics with the process of IML as defined in \autoref{sec:DefIML} on a formal level. The distinguishing between population, sample dataset and model is an inherently statistical view we already adopted in our formalization. IML in the classical sense interprets the model as given and often sets the sample dataset equal to the population, ignoring its representativeness. While IML applies queries to explain a model's prediction, statistics applies models to explain the data. Both processes evaluate or test their explanations, albeit through different procedures. The same applies to the human interpretation: While statistics has defined procedures on how to interpret and present the results (also in terms of scientific writing and reproducibility) \citep{Munafo2017}, IML is just at the beginning and procedures differ from author to author.\\

On a methodological level, statistical models are generally seen as useful and mainly tested for their faithfulness. Statistical metrics which rudimentary consider usefulness are information criteria such as the AIC \citep{Akaike_1974_AIC} or BIC \citep{Schwarz_1978_BIC}. Statistical tests also evaluate faithfulness through confidence and significance statements, which are largely ignored in the evaluation of queries in IML. In addition, statistical models often incorporate assumptions about the distribution of the estimated parameters, which also serve as the query part of the model, and rely, e.g., on the law of large numbers or independence assumptions. These assumptions have to be tested and limit the expressiveness of an explanation. For example, coefficients in a linear regression model estimated via least squares can at best be unbiased and make conclusions about linear relationships (Gauß-Markov Theorem) \citep{McCulloch2001}. However, in comparison, IML methods are also often based on axioms which could be compared to such assumptions.\\

We observe that on a formal level both processes are relatively similar and only differ in how they define correlation and the object they want to explain. On a more methodological level, statistics has already defined several mechanisms for testing and human interaction to vouch for trustworthy results, but its methods lack capacity, making them unusable in an IML setting with normally complex high-dimensional data. Based on this comparison we conclude that we can adapt the statistical process for IML but by means of machine learning models and IML methods to fulfil queries. This formalization of IML does not change the process defined in \autoref{fig1} as it already incorporates the statistical view but extends the formalization of how we handle every stage in the process, by adopting the statistical view.

\section{Elaborated Open Questions in IML}

We formalized IML as a procedure adopted from statistics, but we exchange the tool-set for queries and machine learning models which allow doing statistics on models and complex data, but also have to be evaluated for usefulness. By adopting this statistical view on IML, a considerable amount of possibilities arises as to how we methodologically and formally \textit{do} IML since its application scope is expanded and methods have to be adapted. We summarize these possibilities within three key questions we elaborated, which we argue are crucial for the success and adoption of IML in real-world settings and should give a suggestion in which direction IML should progress. While the first two questions address the formally motivated process of modelling (Q1) and testing (Q2), the last question is motivated by statistical methodology (Q3).\\

\textbf{Q1: What do we want to explain?} \quad Current application of queries focuses on explaining the prediction of a model, ignoring the fact that it is only an extracted relationship from the underlying data. Vice versa, statistics is only interested in explaining the underlying dataset or making inferences about the population, by means of a model but is uninterested in the model itself. We argue that both views have to be combined and IML is a process to either explain model behaviour or discover knowledge from data. \autoref{fig1} shows that we can walk the process backwards, explaining the model, but also infer from the model to the dataset and from the dataset infer back to the population. Inferring backwards from sample data to population is a common procedure in statistics, but for IML we have to place the model upstream in the process since we either assume that our data is too complex to infer directly from it or we are only interested in explaining the model itself. As further one wants to go backwards in the process, as more complex it gets to retrieve trustworthy explanations.\\

When only explaining the model, the object of interest is fixed and all explanations are conditioned on the underlying dataset. But even in this case the question of what we want to explain applies: current queries focus on explaining feature importance for prediction, but they ignore other questions important to the understanding of the model, e.g., explaining uncertainty or explaining the effect of prior architectural choices. But also feature importance can be used for different tasks when explaining the model, such as qualitative OOD evaluation where otherwise test sets containing distribution shifts are required. Particularly in clinical diagnostic distribution shifts are often either unknown or comprehensive OOD test data is lacking.\\

Going a second step backwards, explaining the dataset and only being interested in statements which hold for this sample, can be characterized as doing descriptive statistics. Since we assume that the dataset is of a very high dimension, we have to use a model as a prior step to explanation. By using the model as an extractor of a relationship, the choice and fitting of the model are not trivial anymore. Different models extract different relationships from the data, possibly biasing an interpretation of the dataset. Further, when only summarizing the dataset, one could argue that overfitting is desirable since we want the model as perfectly as possible to fit to the training distribution. However, overfitting also involves the risk of overfitting to a spurious correlation, extracting the wrong correlation, and ultimately leading to a biased, if not \textit{false}, causal statement.\\

Since statements which hold only for a specific sample dataset are only of limited use, general statements which hold for the whole population are much more valuable. To achieve statements which hold as general as possible in addition to the model, the definition of the population and sample procedure of the dataset is not trivial anymore. Biases in the dataset can be partly neglected by having an adequately generalizing model: to make general statements, one needs to explain a generalizing model. Further, for inference to the population, local explanations explaining only one observation in the sample dataset are not sufficient anymore. Global explanations can be achieved by either explicitly global methods such as CAV or, in a frequentist manner, by averaging over several local explanations. Averaging local explanations is not trivial to achieve in computer vision with pixel-level feature importance, since the same local feature can have different spatial positions.\\

\textbf{Q2: How do we evaluate explanations?} \quad Currently, the evaluation of queries is divided into either evaluating the faithfulness of a method, e.g., via the attribution of importance to the model or evaluating the usefulness through application by a regular or domain expert human (e.g. physician). We argue that for a fair, standardized, and comparable evaluation, the resulting assessment has to be quantitative. Note, that this does not rule out human-based experiments, but implies the quantitative evaluation of them. Statistics developed a large repertoire of quantitative metrics, confidence and testing methods for validation, which can be adapted to be applied to queries.\\

Metrics are an informal measure of different aspects of a model or method. Currently, proposed metrics for explanatory methods such as Infidelity or Sensitivity \citep{yeh_fidelity_2019} focus on the evaluation through the importance of an attributed feature to the predicted output accuracy as a proxy measure for a faithful explanation. Relying on metrics which measure the same proxy behaviour to evaluate faithfulness can be dangerous since research in the area of metrics has already shown how volatile rankings of methods are depending on the selected metric for evaluation \citep{Reinke_2021_metrics}. It would be unreasonable to believe that this was also not the case for IML. Further, metrics evaluating faithfulness have to be adapted to also measure global explanations and must apply to different query classes. The usefulness of a model in statistics, but also in other sciences, is commonly evaluated by its complexity, primarily measured through employing information theory. The first work in this area for interpretable models was done by \citet{barcelo_model_2020}. Metrics measuring different aspects of a method can serve as the basis for hypothesis testing and also express the confidence of an explanation.\\

Confidence measures how robust an explanation is. Defined in a statistical sense, it is often expressed as a confidence interval, band, or region, which is a range of estimates for an unknown parameter. In the machine learning sense, confidence is more defined as a score, indicating an event which has a high probability. Achieving small confidence intervals or high confidence scores is the first step to enabling trust in an explanation. But the question remains of how can we express confidence from a query? Probabilistic queries could inherently make statements about their confidence, whereas confidence intervals can be estimated based on sampling and distributional assumptions. Otherwise, data-based experiments can evaluate the confidence of an explanation through, e.g., image transformations or style transfers.\\

At last, testing is applied to verify if an explanation supports a hypothesis sufficient. A hypothesis can hereby be a comparison between explanations or about a single explanation itself. When testing a single explanation, the alternative hypothesis can be either a random explanation (is this explanation more significant than randomness?) or a specifically selected explanation (does this explanation attribute significant more to the left lung than to the right lung in the CT image?). The definition of significance is not trivial, as it is also a controversial subject in statistics, and introductions of p-values should be dealt with caution as they tempt to be misused. Further, we could not only test hypotheses about the explanation itself but also if implied assumptions hold and underlying axioms are sufficient. Testing and cross-validating between explanations differ in complexity depending on if one wants to compare methods within one query class or between query classes. While within a class comparing two explanations is less difficult since the same type of explanation is present, between-class explanations have to be tested based on comparable metrics or summarizing meta information.\\

\textbf{Q3: How do we conduct the application and development of ill-equipped methods?} \quad As other research stressed before, queries do not have to be perfect to be trustworthy \citep{doshi-velez_towards_2017}. But it has to be communicated what their capabilities are and how these limit the explanation-based causal statement. This is of cause an age-old story in science, linear models can only extract linear relationships, and it has to be asserted for what kind of data this is enough and where relationships are more complex. Thus, future frameworks and research have to evaluate under what circumstances certain query classes fail or succeed, and match them with modalities, domains, and specific questions they answer. Their adaptation to, e.g., volumetric and video data should be encouraged, which has been largely ignored at this point, and examined if current methods also prevail on these modalities. Further, the development of new methods has to be tied to quantitative evaluation, and not based on a qualitative assessment which favours the confirmation bias, leading to the application of unfaithful methods.\\

While most statistical methods are based on distributional assumptions, deep learning methods are only based on a smaller set of weaker assumptions, which is one of the key advantages in performance over statistical models. But the inclusion of strong assumptions as inductive biases for queries can also have advantages by making them easier to understand and verify. On the contrary, these biases can weaken their capacity and bias the explanation.\\

The selection of a good explanation depends on the underlying model, data, and question, which can be very time-consuming and knowledge depending. At the moment, explanatory methods are applied by stakeholders in a non-systematic and swift way, only to get an intuition on the behaviour and ignoring to challenge the explanation. While this is not inherently dangerous, wrong interpretations based on biased explanations can still harm the application in the long run. Frameworks give a theoretical basis to select and evaluate explanations depending on all these dependencies, but it is non-trivial and time-consuming to implement them rigorously. Auto IML methods can overcome this challenge and automate procedures determined by frameworks. They would be easy to use, adoption friendly, and minimize the risk of a wrong explanation. In addition, they neglect possible biases induced by qualitative assessment and confirmation biases when selecting an explanation. Undoubtedly, this approach only works if quantification, comparability and standardized evaluation between methods have been established, as already promoted in the previous questions.\\

\section{Conclusion}

To conclude, we provided evidence for the need for XAI, its lack of formalization, and its lagging adaption in the application. Further, we elaborate on conceptual similarities between IML and classical statistics. Based on these similarities, we present a formalization of IML along the lines of a statistical process. At last, we inferred three key questions and their implications, when adopting the statistical view on IML. By transferring the statistical procedures and methods to IML, IML can become a much more formalized area of research, with defined and tested practices, all with the premise of enabling and verifying trustworthy results. Through our proposed view of the field, the defined scope of IML is expanded, i.e. going beyond explaining predictions towards knowledge discovery from data analogously to a classical statistical setting, thereby closing the gap between correlation and causation.\\

Of course, simply putting on the \textit{statistics glasses} will not solve the especially mythological problems of IML. Instead, our work aims to spark a discussion based on the hypothesis that IML should be considered as a statistical process employing machine learning models and query methods, which begs question: \textit{``have we been doing classical statistics all along?''}.

\section*{Ethical Statement}

There are no ethical issues.

\section*{Acknowledgments}

Part of this work was funded by the Helmholtz Imaging (HI), a platform of the Helmholtz Incubator on Information and Data Science.

\bibliographystyle{named}
\bibliography{references, references_digital}

\end{document}